\def\year{2019}\relax
\documentclass[letterpaper]{article} 
\usepackage{aaai19}  
\usepackage{times}  
\usepackage{helvet}  
\usepackage{courier}  
\usepackage{url}  

\usepackage{enumerate,paralist}
\usepackage{subfig}
\usepackage{graphicx}

\usepackage{booktabs}
\usepackage{amsfonts}
\usepackage{multirow}
\usepackage{comment}

\usepackage{xspace}
\usepackage{xcolor}
\usepackage{enumitem,amsmath}

\usepackage{CJKutf8}

\usepackage{graphicx}  
\begin{document}
%
\title{Entropy-Enhanced Multimodal Attention Model\\ for Scene-Aware Dialogue Generation}

\author{
    \begin{tabular}[c]{@{}c@{}}
    Kuan-Yen Lin$^{\dag\star}$\quad
    Chao-Chun Hsu$^{\dag}$\quad
    Yun-Nung Chen$^{\star}$\quad
    Lun-Wei Ku$^{\dag}$\quad
    \end{tabular}\\
    $^\dag$Academia Sinica, Taiwan
    $^\star$National Taiwan University, Taiwan\\
    \texttt{b03612017@ntu.edu.tw}\quad
    \texttt{joe32140@iis.sinica.edu.tw}\\
    \texttt{y.v.chen@ieee.org}\quad
    \texttt{lwku@iis.sinica.edu.tw}\quad
}

\maketitle
\setlength{\parindent}{1em}
\begin{abstract}
With increasing information from social media, there are more and more videos available.
Therefore, the ability to reason on a video is important and deserves to be discussed.
The Dialog System Technology Challenge (DSTC7) \cite{DSTC7} proposed an Audio Visual Scene-aware Dialog (AVSD) task, which contains five modalities including video, dialogue history, summary, and caption, as a scene-aware environment.
In this paper, we propose the entropy-enhanced dynamic memory network (DMN) to effectively model video modality.
The attention-based GRU in the proposed model can improve the model's ability to comprehend and memorize sequential information.
The entropy mechanism can control the attention distribution higher, so each to-be-answered question can focus more specifically on a small set of video segments.
After the entropy-enhanced DMN secures the video context, we apply an attention model that incorporates summary and caption to generate an accurate answer given the question about the video.
In the official evaluation, our system can achieve improved performance against the released baseline model for both subjective and objective evaluation metrics.

\end{abstract}
\begin{CJK*}{UTF8}{bsmi}

\section{Introduction}
Recently a lot of attention has been put to visual question answering (VQA), because many real-time problems also lie in the category of VQA. 
The most well-known is the VQA challenge, which is a huge dataset that contains lots of visual questions and their answers. In VQA, the system needs to correctly answer the question with response to a given image. The VQA task requires a model equipped with the ability of natural language understanding and visual reasoning. 


With images wildly exploited in all those tasks, the spotlight gradually transfers to videos such as video captioning and video question answering. Videos are comprised of many images along with the audio information which provides the model with listening capability. The visual part contains the information in a frame and information between frames. In recent days, we can easily access video-related datasets.  Videos can open more potential research topics for the research community compared with image,
Video tasks require an ability to reason over the given video; however, the ability to understand the whole video is very challenging. Thus, in Audio Visual Scene-aware Dialog(AVSD) \cite{hori2018end}, the task will be given with multiple modalities and expect to gain a better understanding for the video.

The scenario setting of the task is that for a video there are a summary, a caption, and a dialogue that are all related to the video. The caption is a description of the video. As for the summary, it is like a conclusion about the dialogue. A system of this task has to generate a response for the last question in the dialogue and the previous question-and-answer (QA) pairs will be regarded as the dialogue history. With all the above, additional usable sources aside from video, the system is expected to generate a natural language response that is accurate according to the video contexts. 

In our work, we extensively use Attention mechanism \cite{bahdanau2014neural}. Since the debut of attention mechanism, it has shown to be useful in lots of fields of study. Attention helps the model to denoise and focus on a rather important part in sources. The provided captions and summaries could be useful in some questions; additionally, texts are less noisy than videos, so Attention can effectively extract the essence in them. Video requires a stronger technique to denoise to acquire useful content. Because videos are lengthy and very informative, we can view it as the most important modality. We use DMN on videos to strengthen the memorizing ability of the model. The attention-based GRU allow DMN to use positional information in a video more efficaciously. However, the differences between frames in the video are hard to detect, so the distribution of attention-based GRU may be sparse. Hence, we add entropy to the loss to enhance our DMN, which we will call it Entropy-enhanced DMN.  

\section{Related Work}
Our model is built on top of Dynamic Memory Network (DMN) \cite{kumar2016ask,xiong2016dynamic}.
Therefore, we first introduce DMN, discuss visual question answering, which is highly related to this task, and review some video-related tasks.

\subsection{Dynamic Memory Network}

Attention mechanism is commonly used in nowadays deep learning tasks \cite{xu2016ask,xu2015show,luong2015effective} due to its superior performance in many tasks.
There are two types of mechanisms, hard attention and soft attention.
The hard attention is inspired by the REINFORCE \cite{williams1992simple} algorithm.
If there is an object in an image to detect, hard attention needs to locate the exact edge of an object in a given image.
The soft attention does not need to acquire the exact edges of the object but a relative position.
Because the soft attention is not discrete, it allows models to be trained in an end-to-end fashion. 

DMN is widely used in question answering tasks, which is composed of four parts, an \emph{input module}, a \emph{question module}, an \emph{episodic memory module} and an \emph{answer module}. 
The input module encodes the input data, called input facts.
The question module encodes the question into the question embedding.
The episodic memory module needs to retrieve the answer from the sources given the question.
The answer module then uses the information from the episodic memory module to generate the answer.
The episodic memory module allows DMN to read and remember sequential information.
The attention-based GRU in the episodic memory module gives DMN a better use of ordering and positional information in a video compared to the basic attention. Therefore, it is reasonable to apply DMN instead of the basic attention to videos.

\subsection{Visual Question Answering}


Lots of spotlights have been cast on visual question answering(abbreviated as VQA) since many of the real-life problems lie under the category of VQA. In the task, there will be a given image, and a question will be based on the image. To reply a correct response, not only does the model needs to have the ability to reason on a given image but it has to understand the natural language question. Since we can view a video as a composition of multiple images, video QA is similar to VQA problem. VQA challenge \cite{antol2015vqa} is a famous visual question answering challenge. Last year the challenge champion \cite{anderson2017bottom}, they mimicked the procedure of humans solving VQA problems. Most of the time humans will listen to the question and use the question as a guide to finding useful parts in the image. With the scenario setting, they exploited Faster R-CNN \cite{ren2015faster} to extract image feature. Faster R-CNN is a robust model that can capture objects in images. The step is called the Bottom-up mechanism. With this action, the model will first choose top k objects that are likely to be the key answer to the question. As for the top-down mechanism, the top k candidates and the question will be used to compute the attention distribution and in the end use for generating the answer.

\subsection{Video Related Tasks}




With deploying CNN \cite{lecun2015deep} in a model, lots of image concerning tasks have shown great promise. Therefore, the spotlight has gradually moved to video concerning tasks. C3D \cite{tran2015learning} expands 2D-CNN into 3D-CNN. Though it is straightforward, 3D-CNN is hard to train and could not share the benefits from pre-trained Imagenet. Afterward, I3D \cite{carreira2017quo} manages to use pre-trained Imagenet by bootstrapping from 2D filters to 3D filters. Both image features and optical-flow features will be used in the process. \\

Recently, a lot of focus has been on video in the computer vision area. Tasks such as video classification, video question answering, and video captioning are hot topics. In \cite{wang2018bidirectional}, they are generating suitable captions for the video. They make good use of bidirectional LSTM \cite{hochreiter1997long}, which the forward path represented the history and the backward path meant to be the future. Each path will be used to predict events and afterward, they will fuse the predicted events to generate a caption. In \cite{gao2018motion}, they also built on Dynamic Memory Network to perform on Video Question Answering. Moreover, they also use conv-deconv network to generate facts in DMN. Lastly, they use an ensemble method in the model. AVSD is a video dialogue generation task, so comparing with general video QA tasks, the generated answer need to be based on a dialogue history. Our model enhances DMN with entropy, which is a way to help the model differentiate the differences between frames.

\section{Data}
Before we dive into our method, we will first elaborate the date format. In each round, as shown in Table \ref{table:dataset}, the model is given with five modalities, a dialogue history, a caption, a summary, and a video. Dialogue history is an interactive conversation of the questioner and an oracle about the video. After they finished their dialogue, the summary will be written by the questioner. The caption is a sentence that describes the content of the video. The system is given with all of the above sources and a question about the details in the video to generate an appropriate answer. The official dataset contains 7,659 of training data, 1,787 of validation data, and 1,710 of testing data.

\section{Proposed Approach}
\begin{figure*}[t!]
  \centering
  \subfloat[Episodic Memory Module]{\raisebox{-0ex}{ \includegraphics[height=0.25\textheight,width=0.49\textwidth]{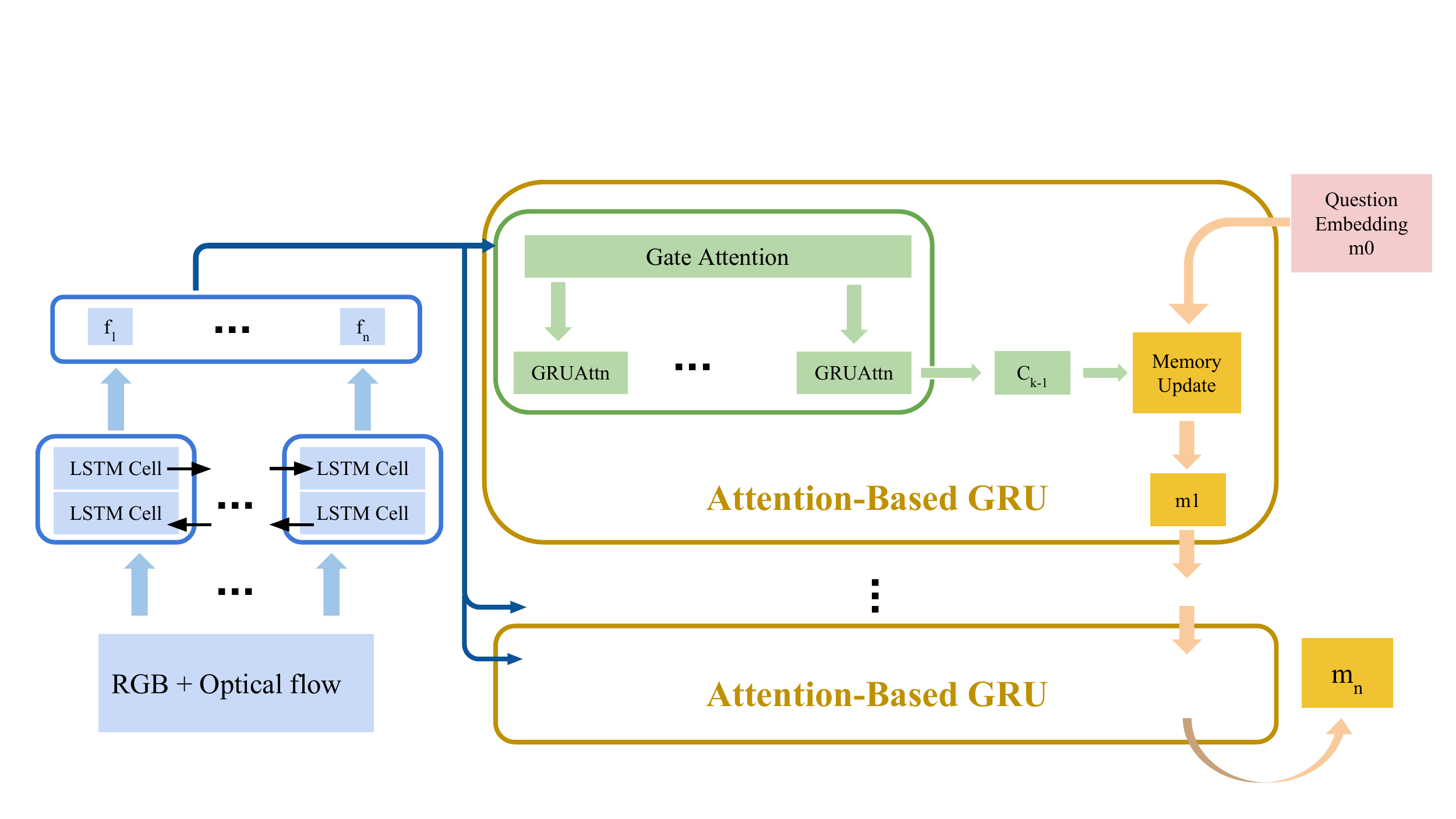}}\label{fig:f1}}
  \hfill
  \subfloat[Overall Model Structure]{\includegraphics[height=0.25\textheight,width=0.49\textwidth]{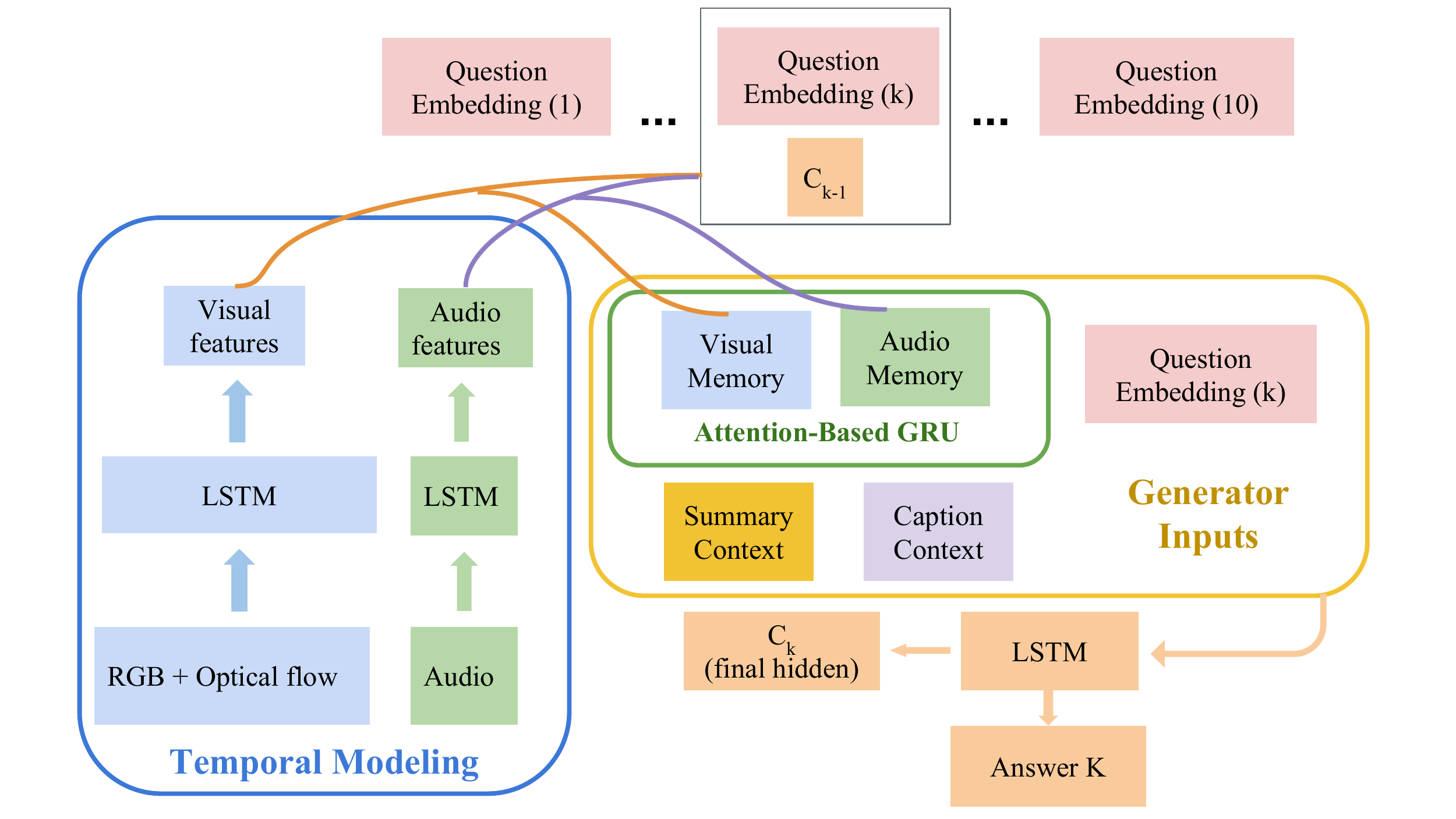}\label{fig:f2}}
  \caption{Each episodic module will be used to retrieve useful information in the input facts. In (a), \textsl{\small RGB + Optical flow} means the combination of RGB and Optical flow features. RGB and Optical flow features come from I3D model. (b) is the overall architecture of our system. Audio features and visual features will use different episodic memories to generate visual memory and audio memory. Attention mechanism will use a question as a query to generate a summary context and a caption context.
  }
  \label{attention_visualization}
  \vspace{-1em}
\end{figure*}

The input of our model will be a summary, a caption, a dialogue history, and a video. The video will be separated into 2 parts, audio, and visual part, which each part will have respective feature. Two different networks will extract audio and visual features. Our entropy-enhanced DMN will encode the extracted visual features, and so will the audio features. Attention will encode summary and caption separately and generate summary context and caption context. The encoded modalities will use while decoding. The detail of each component in the model is described below.
\subsection{Video Feature Extraction}
Each video will be separate into visual part and audio part. The audio features are extracted by Vggish \cite{hershey2017cnn}. The visual part will be extracted by Two-Stream Inflated 3D ConvNets (I3D) \cite{carreira2017quo}.
\subsubsection{Visual Feature}
I3D is introduced to extract video's visual features. The "Two-stream" of the model indicates the separation of video into two part to acquire features. Videos are composed of several frames. There will be RGB information in each frame and optical flow information between frames. These are the two streams, optical flow stream, and RGB stream. I3D model highlights to use the pre-train ImageNet. Each filter in I3D model will inflate the 2D filter into 3D, and the parameters are shared by the pre-trained ImageNet. 

\subsubsection{Audio Feature}
The Vggish model extracts the audio part of the video. The Vggish uses the same architecture as the well-known Vgg model and replaces the input image with an audio spectrogram.

\subsection{Soft-Attention for Text}
In AVSD, the dataset contains additional caption and summary regarding the video. The caption is a sentence of the description of the video. As for the summary, it is generated after the dialogue history, so it summarizes the video with the knowledge of dialogue history. Both of the caption and summary may contain keywords or key information for answering the question. Therefore, we apply soft-attention on both modalities. In this paper, we will refer soft-attention to Basic Attention.
\begin{equation}
\begin{aligned}
  \label{eq:softmax}
\text{attn}_{1} &=  {W}{t} + {b}\\
\text{attn}_{2} &=  {W}{q} + {b}\\
{e} &= {W}\tanh(\text{attn}_{1} +\text{attn}_{2}) + {b}\\
{\alpha} &= \frac{\exp({e})}{\sum^{l}_j \exp({e}_{j})},
\end{aligned}
\end{equation}
\textsl{$t$} represents text which can be caption or summary. \textsl{$q$} is the question embedding. \textsl{$l$} means the length of the text.

\subsection{Dynamic Memory Network for Video}
Dynamic Memory Network (DMN)\cite{kumar2016ask} shows excellent results in previous QA, VQA, and some video QA. Because of the episodic memory module, the model will have a better ability to remember retrieved segments in the video. Moreover, in each episodic memory, it will focus on different part of the video, so the model can respond better while facing a more complex sentence. Thus, DMN is also integrated into our model to improve the system's ability to memorize the content of visual and audio information. We separate our DMN into input module, episodic module, and answer module to further explain our model.

\subsection{Input Module}
In our task, the dialogue history contains several QA pairs and a video, and both of them could play important roles in responding to the question. We treat the QA pairs as additional data for training our model.

In the input module, the question to be answered will be transformed into an embedding. We use Bidirectional LSTM to encode our question, which we concatenate the cell states of forwarding LSTM and backward LSTM as the question embedding.

As for the extracted visual and audio features of a video, we also use LSTM to generate two kinds of embedding. Since both of them are spatial-temporal features, LSTM can strengthen the global information of the whole features and filter out some noise in both modalities. In general DMN, this encoded embeddings are named as \textsl{input facts}, \textsl{F} =  [$\textsl{f}_{1},...,  \textsl{f}_{N}$]. N is the number of frames.

\subsection{Attention-based GRU}
Before we dive into how the Episodic Memory Module works, we will illustrate Attention-based GRU \cite{kumar2016ask}.
General attention \cite{bahdanau2014neural} mechanism retrieves information for answering concerning the input query. Although the concept is intuitively straightforward, soft-attention is devoid of the ability to ordering and position awareness. 
Attention-based GRU replaces the update gate from original GRU with the attention gate. 
\begin{equation}
{h}^{i} = {g}^{i}_{t} \circ {h}^{i}_{'} + (1 - {g}^{i}_{t}) \circ {h}^{i-1}
\end{equation}
\textsl{${g}^{i}_{t}$} denotes the attention gate. \textsl{$\circ$} represents Hadamard product operation. \textsl{$t$} means t-th episodic memory. \textsl{$i$} indicates i-th Attention-based GRU cell in the Attention-based GRU.

Each Attention-based GRU cell will consider question embedding \textsl{q} and previous generated memory \textsl{${m}_{t-1}$} as guidance to attend on important part of \textsl{input facts}.
\begin{equation}
{z}^{i}_{t} = [{f}_{t} \circ q ; {f}_{t} \circ {m}_{t-1}]
\end{equation}
\begin{equation}
{Z}^{i}_{t} = {W}^{(2)} tanh( {W}^{(1)}{z}^{i}_{t} + {b}^{(1)}) +  {b}^{(2)}
\end{equation}
\begin{equation}
  \label{eq:softmax}
  {g}^{i}_{t} = \frac{\exp({Z}^{i}_{t})}{\sum^{N}_j \exp({Z}^{i j}_{t})},
\end{equation}
\textsl{$;$} means concatenation. \subsection{Episodic Memory Module}
With an episodic memory module, the system are more capable of redeeming information from the \textsl{input facts} while it faces a rather complex question. In a more complex question scenario, the question may have multiple information needed to be retrieve from the sources. With the usage of Attention-based GRU, episodic module will possess the aptness of ordering and positional information. 

In episodic memory module, it will update M times (M episodes, M is a hyperparameter). In each updating episode, inside an episodic memory module, \textsl{input facts}, \textsl{F} =  [$\textsl{f}_{1},...,  \textsl{f}_{N}$], and question embedding \textsl{q} would generate an attention gate for this episode. The attention gate will be used in Attention-based GRU, which will traverse through \textsl{Input facts}. After the traversal of the \textsl{input facts}, Attention-based GRU can output a \textsl{contextual hidden state} ${c}_{t}$. The contextual hidden state ${c}_{t}$ will be used as the updated memory ${m}_{t}$ in this episode.
\begin{equation}
{c}_{t} = [q ; {h} ; {m}_{t-1}]
\end{equation}
\begin{equation}
{m}_{t} = Relu( {W}{c}_{t} + {b}) 
\end{equation}

\subsection{Multimodal Fusion}
In this task, the system will have video, dialogue history, caption, and summary. Each of the above is a modality that could contribute to better answering. We acquire textual context embedding from summary and caption by Basic Attention. Afterward, DMN will be applied to audio and visual features to generate visual context embedding and audio context embedding. 
\begin{equation}
  \label{eq:softmax}
{v} = \sum^{m}_{j=1} \beta_{j}{C}_{j}, \beta_{n} = \frac{\exp({C}_{n})}{\sum^{m}_{j=1} \exp({C}_{j})},
\end{equation}
where ${v}$ denotes final context encoding. ${C}_{n}$ means the different context embedding from different modalities. With question as guide, the gating mechanism can select which modalities are more important for answering the question.

\subsection{Dialogue History Modeling}
Each question in dialogue history will be used to generate the answer. When decoding answer, the decoder will also consider the hidden state of the previous answer. Since each question in QA pairs and the question to be answered are all based on the previous question, we hypothesize that it is crucial to let the model knows the previous answer while decoding.

\begin{equation}
{s}_{t}^{n} = LSTM ([{s}_{T}^{n-1};{v};{y}_{t-1}^{n}])
\end{equation}
\begin{equation}
P({y}_{n} | {y}_{1}, ..., {y}_{t-1}) = \textit{Softmax}({W}{s}_{t}^{n} + {b})
\end{equation}
$n$ denotes as the number of the question that the model is currently answering. $t$ denotes timestamp. ${s}_{T}^{n-1}$ means the previous generated answer's last hidden state. After obtaining the word dictionary distribution, we use beam search to produce the answer. 

\subsection{Entropy}
At first, the distribution of DMN is very sparse. Thus, we add an entropy in our loss function to make our distribution more concentrated. We call it Entropy-Enhanced DMN.

\begin{equation}
Loss = - \frac{1}{K}\sum_{i}^{K} p(y_{i})\log p(\overline{y}_{i}) - \gamma\sum_{m}\sum_{r} p(r) \log p(r)
\end{equation}
$m$ denotes audio and visual modalites. $r$ represents each distribution in $m$. $\gamma$ is $a$ hyperparameter and $K$ denotes sentence length.

\section{Experiments}
\begin{table*}[htbp]
\centering
\small
\resizebox{\textwidth}{!}{
\begin{tabular}{@{}lccccccccccc@{}}
\toprule
            & BLEU-1 & BLEU-2 & BLEU-3 & BLEU-4 & METEOR & ROUGE-L & CIDEr \\ \midrule
Baseline \cite{hori2018end} &   0.273  &0.173 &0.118 & 0.084 & 0.117 &0.291  &0.766 \\ 
 \midrule
Basic Attention (D)     & 0.317  & 0.185    & 0.121 & 0.083    & 0.132  & 0.330 & 0.760 \\ 
Basic Attention (D, S, C)    & 0.318  & 0.183    & 0.120 & 0.083    & 0.122  & 0.318    & 0.765 
\\
 \midrule
DMN (A, V, D, S, C) & 0.316  & 0.187    & 0.127 & 0.089 & 0.123  & 0.327 & 0.821 
\\ 
Entropy-Enhanced DMN (A, V, D)    & 0.320  & 0.185  & 0.121 & 0.083 & 0.121  &  0.319  & 0.736 \\ 
Entropy-Enhanced DMN (A, D, S, C)     & 0.329  & 0.194   & 0.130 & 0.093    & 0.127  & 0.330 & 0.858 \\ 
Entropy-Enhanced DMN (V, D, S, C)    & 0.323  & 0.184 & 0.119 & 0.081  &  0.123  & 0.321 & 0.707 \\ 
Entropy-Enhanced DMN (A, V, D, S, C, no d.m.) & \textbf{0.333} & 0.194    & 0.128 & 0.091    & 0.127  & 0.327 & 0.780
\\
Entropy-Enhanced DMN (A, V, D, S, C) & 0.331  & \textbf{0.196}    & 0.130 & 0.091 & 0.128  & 0.333 & 0.843 
\\ 
Entropy-Enhanced DMN (A, V, D, S, C, $M = 3$)   & 0.329  & 0.195   & \textbf{0.131} & \textbf{0.093} & \textbf{0.129}  & \textbf{0.334} & \textbf{0.880}
\\
\bottomrule

\end{tabular}
}
\vspace{-.5pc}
\caption{\small The first part is the official objective evaluation values. The second part contain two subsection and both of them don't have audio or visual information. In the third part, each subsection will all have at least one video features. DMN represents the general DMN, and Entropy-Enhanced DMN will be our proposed model. Each DMN will have $M=2$: two updating times in a DMN. In the final row of the third part, $M=3$ means we increase to three times for DMN to update. The capital letter in the parentheses means the given modality. (D: dialogue history; S: summary; C: caption, A: audio, V: visual)} 
\label{table:official}
\end{table*}

\begin{table*}[htbp]
\centering
\small
\resizebox{\textwidth}{!}{
\begin{tabular}{@{}lccccccccccc@{}}
\toprule
& BLEU-1 & BLEU-2 & BLEU-3 & BLEU-4 & METEOR & ROUGE-L & CIDEr & Human \\ \midrule
Baseline \cite{hori2018end}   & 0.626  & 0.485   & 0.383 & 0.309 & 0.215  & 0.487 & 0.746 & 2.848

\\ \midrule
Entropy-Enhanced DMN (A, V, D, S, C)   & \textbf{0.641}  & \textbf{0.493}   & \textbf{0.388} & \textbf{0.310} & \textbf{0.241}  & \textbf{0.527} & \textbf{0.912} &
\textbf{3.048}
\\ \bottomrule

\end{tabular}
}
\vspace{-.5pc}
\caption{This table is the final result released by the official.}
\label{table:dataset}
\vspace{-1pc}
\end{table*}

\subsection{Experimental Setups}
 We separate modalities into 2 main parts, which are Video and Text. Experiment wishes to show us how two different kinds of modalities would affect our system. We also conduct several trails to test episodic memories. Lastly, we will test our claim of questions underlying connection.
 
 In our submitted system, I3D extracted RGB features and optical flow features are combined as a visual modality, since we found out that the separation of two visual features will deteriorate the performance of our system. Both audio and visual features will use dynamic memory network to generate embeddings. Both of DMN is set to 2 episodic memory ($M = 2$). All of our modalities' hidden size is set to 128. We use Glove \cite{pennington2014glove} to initialize our word embedding. We test all experiments on the prototype testing set.
 %

\subsection{Dialogue History, Caption, Summary}
Comparing text modalities with video modalities are rather clean. In this experiment, we are curious how the system will perform when excluding video. In Table~\ref{table:official}, the result of this setting (\textsl{Basic Attention (D, S, C)}) shows that it is not as comparable as our final submitted system (\textsl{Entropy-Enhanced DMN (A, V, D, S, C)}). Though videos are noisy and sometimes dirty, they contain important information that caption and summary don't have. Thus, text-based modalities could be less useful. 

\subsection{Dialogue History and Dialogue Modeling Analysis}
Dialogue history may be the most informative modality in this task. In the dialogue history, many questions may be in condition on previous questions. Since they are bonded, considering previous answers will be helpful for answering current questions. Thus, we hypothesize that if the model has the prior information, the system will be more qualified to generate a more suitable response, which we will call this hypothesis: dialogue modeling. In this section, we will first experiment on only giving the model dialogue information. Thereafter, we will explore whether our dialogue modeling is useful.

Since dialogue history is such an important source to answer a question, we wonder how our system will perform if it only has the information of dialogue history. In Table~\ref{table:official}, \textsl{Basic Attention (D)} row shows the final result. All values aren't comparable to our submitted system (\textsl{Entropy-Enhanced DMN (A, V, D, S, C)}). Moreover, our system couldn't respond in a scene-aware style because it is restricted tosingle modality.

As for dialogue modeling, an essential setting in our system. We preclude the previous answer's last hidden state to be presented in the model; thus, the model won't have the information on previous questions. In Table~\ref{table:official}, \textsl{Entropy-Enhanced DMN (A, V, D, S, C, no d.m.)} row is the final result. Though BLEU-1 is slightly higher than our submitted system (\textsl{Entropy-Enhanced DMN (A, V, D, S, C)}), other values aren't comparable.

\subsection{Video Analysis}
Videos contain the key to answer questions. We first experiment on giving our model video information, so the system has audio and visual information. Thereafter, we separate audio and visual features to test each modality.

We can see \textsl{Entropy-Enhanced DMN (A, V, D)} performs terribly in the final result. Therefore, summary and caption will help our system to generate better responses. We can know the audio features could be less noisy than the visual features, since \textsl{Entropy-Enhanced DMN (A, D, S, C)} performs better than \textsl{Entropy-Enhanced DMN (V, D, S, C)}.

\begin{table*}[h!]
\centering
\small
\begin{tabular}{llll}
\toprule
\multirow{16}{*}{\begin{tabular}[c]{@{}l@{}}Dialogue\\ History\end{tabular}} & \multicolumn{1}{c}{Question}                                 & \multicolumn{1}{c}{Answer} &  \\ \cmidrule{2-4} 
& \textbf{\#1} does she enter the room or was she already there ?           & \begin{tabular}[c]{@{}l@{}}she is there, i think she starts the camera and then walks \\ toward the fridge .\end{tabular} &  \\ 
\cmidrule{2-4} 
& \textbf{\#2} does she get anything out of  the fridge ?                   & she gets out a carton of something . &  \\ \cmidrule{2-4} 
& \textbf{\#3} does  she say anything ?                                     & no , she is silent the entire time .                                              
&  \\ \cmidrule{2-4} 
& \textbf{\#4} does  she get a cup ? & the glass is already on the counter .   
&  \\ \cmidrule{2-4} 
& \textbf{\#5} does   she eat or drink anything ?                           & \begin{tabular}[c]{@{}l@{}}she  takes a sip of the drink while walking back to her\\ computer . \end{tabular}                                                                     &  \\ \cmidrule{2-4} 
                                                                             & \textbf{\#6} does she pour what was in the carton into something ?        & yes she pours a tiny amount  into the glass .                                                                                           &  \\ \cmidrule{2-4} 
                                                                             & \textbf{\#7} does she put the carton back in the fridge then ?            & yes , she puts it back and then picks up the cup .                                                                                      &  \\ \cmidrule{2-4} 
                                                                             & \textbf{\#8} does she smile or have any emotion ?                         & \begin{tabular}[c]{@{}l@{}}no , but she does look back at the camera a \\few times .\end{tabular}                                                                                 &  \\ \cmidrule{2-4} 
                                                                             &\textbf{\#9} \begin{tabular}[c]{@{}l@{}} does she ever mess with the tons of bananas on\\ her counter ?\end{tabular} & \begin{tabular}[c]{@{}l@{}}no , but she does turn the computer / camera\\ around as  she sits down at a table at the end .\end{tabular} &  \\ \midrule
Caption                                                                      & \multicolumn{2}{l}{ \begin{tabular}[c]{@{}l@{}}a women gets a carton out of the fridge , pours some in a glass , and puts the carton back before sitting down .\end{tabular}}                                                                   &  \\\midrule
Summary                                                                      & \multicolumn{2}{l}{one person pours something from a refrigerator , then sits at a desk and starts working .}  \\     \bottomrule
\end{tabular}

\label{table:dialogue_sample}
\caption{A sample of dialogue history, a caption, and a summary of the video}
\vspace{-.3cm}
\end{table*}

\begin{table}[]
\centering
\begin{tabular}{l|l}
\toprule
Question     & \begin{tabular}[c]{@{}l@{}}does that look like a kitchen or a break \\ room ?\end{tabular} \\ \midrule
Ground Truth & it looks like her {\color{blue} kitchen} . \\
Proposed ($M = 3$) & \begin{tabular}[c]{@{}l@{}}it appears to be a  {\color{blue}kitchen} .\end{tabular}                    \\ 
w/o Video    & it looks like a {\color{red}living room }. \\ 
\bottomrule
\end{tabular}
\caption{
\small A sample of the question and ground truth. \textsl{M = 3} is the model with three episodic memory. \textsl{w/o Video} means that the system is only given caption, summary, dialogue history and there is no video. }
\label{table:case_study}
\vspace{-.3cm}
\end{table}

\subsection{Episodic Memory Testing}
In many episodic memories, each episodic memory will focus on different part of the given features. With this trait, while facing a more complicated question, the system has a higher chance to respond correctly.  In this experiment, we will test the number of episodic memory.

Setting the number of episodic memory to two (\textsl{Entropy-Enhanced DMN (A, V, D, S, C)}), the model is successfully trained to converge. Since we know that the more episodic memory would be the better, we wonder when will the performance saturated. Therefore, we set the number of episodic memory to 3. Nevertheless, the outcomes are only slightly better in some categories and the training and testing time drastically increases due to the additional parameters from the third episodic memory.



\subsection{Entropy Analysis}
Aside from the distribution of Attention-based GRU becoming more concentrated. After adding entropy to our loss function, the result is better. We can see in Table~\ref{table:official} \textsl{DMN (A, V, D, S, C)} is not camparable with our submitted system.

\subsection{Official Evaluation}
Table~\ref{table:dataset} is the final result released by the official. Our submitted system outperforms the released baseline model for both subjective and objective evaluation metrics.

\section{Discussion}

\subsection{Training Data Quality}
In the dataset, we discovered some issues about the provided answers, and we address them below.
\begin{compactitem}
    \item  We find that the ground truth is slightly weird which may be hard for the system to learn.
    In the following example, the question asks whether the character is at work? The question requires the system to recognize the environment or behavior of the role in the video. The question is challenging to answer. However, the ground truth answer provides an ambiguous answer that could lead to bas performance. For example, 
    
    \textbf{question}: is he at work ?
    
    \textbf{answer}: hard to say - he is sitting in the hallway by himself .
    
    \item Lots of to-be-answered questions in the training data ask additional information, such as "Is there anything that I missed?". Often, the ground truth is a generic reply, such as "that is all that happens."
    
    \item
    Often caption and summary are useless for answering questions. For instance, in the following case, both caption and summary don't contain useful information to answer the question.
    
    \textbf{question}: does the man shift the pillow?
    
    \textbf{caption}: a person is laughing and snuggling a pillow and taking a sandwich.
    
    \textbf{summary}: a man sitting hugging a pillow
    
\end{compactitem}


\subsection{Qualitative Analysis}
In the case, we can see that the dialogue in Table 2 covers most the content of the video. However, the last question in Table 3 is \textbf{\textsl{does that looks like a kitchen or a break room ?}}. We can see that when the model is provided with video, it generates a more suitable answer. It perfectly answers \textbf{\textsl{it appears to be a kitchen .}} Nonetheless, we can see when the model that doesn't have video information. It replies \textbf{\textsl{it looks like a living room .}} which is obviously a randomly generated answer. Hence, with video information, the system can generate more informative responses.


\section{Conclusion}

In our work, we use multiple modalities to let the model gain a scene-aware ability. Using dialogue history as data augmentation results to be useful and improves the overall performance. Adding previous dialogue history also escalate the overall result. We perform DMN on video to reinforce the model's ability to memorize important information. DMN also helps the system to extract various useful information in the video. 



\end{CJK*}

\bibliographystyle{aaai}
\bibliography{aaai}

\end{document}